\documentclass{article}
\usepackage{ijcai25}
\usepackage{amssymb,amsmath}
\usepackage{epstopdf}
\usepackage{multirow}
\usepackage{color}
\usepackage[table]{xcolor}
\usepackage{colortbl}
\usepackage{array}
\usepackage{arydshln}
\usepackage{bm}

\usepackage{times}
\usepackage{soul}
\usepackage{url}
\usepackage[hidelinks]{hyperref}
\usepackage[utf8]{inputenc}
\usepackage[small]{caption}
\usepackage{graphicx}
\usepackage{amsmath}
\usepackage{amsthm}
\usepackage{booktabs}
\usepackage{algorithm}
\usepackage{algorithmic}
\usepackage[switch]{lineno}

\title{Dual-Perspective United Transformer for Object Segmentation in Optical Remote Sensing Images}

\author{
Yanguang Sun$^1$\and
Jiexi Yan$^2$\and
Jianjun Qian$^1$\and
Chunyan Xu$^{1}$\and
Jian Yang$^{1}$\And
Lei Luo$^{1}$
\thanks{Corresponding author.}\\
\affiliations
$^1$PCA Lab, Nanjing University of Science and Technology, Nanjing, China\\
$^2$School of Computer Science and Technology, Xidian University, Xian, China\\
\emails
Sunyg@njust.edu.cn
}

\begin{document}

\maketitle

\begin{abstract}
Automatically segmenting objects from optical remote sensing images (ORSIs) is an important task. Most existing models are primarily based on either convolutional or Transformer features, each offering distinct advantages. Exploiting both advantages is valuable research, but it presents several challenges, including the heterogeneity between the two types of features, high complexity, and large parameters of the model. However, these issues are often overlooked in existing the ORSIs methods, causing sub-optimal segmentation. For that, we propose a novel Dual-Perspective United Transformer (DPU-Former) with a unique structure designed to simultaneously integrate long-range dependencies and spatial details. In particular, we design the global-local mixed attention, which captures diverse information through two perspectives and introduces a Fourier-space merging strategy to obviate deviations for efficient fusion. Furthermore, we present a gated linear feed-forward network to increase the expressive ability. Additionally, we construct a DPU-Former decoder to aggregate and strength features at different layers. Consequently, the DPU-Former model outperforms the state-of-the-art methods on multiple datasets. Code: \href{https://github.com/CSYSI/DPU-Former}{https://github.com/CSYSI/DPU-Former}.
\end{abstract}

\section{Introduction}
Object segmentation in ORSIs has gained considerable attention \cite{SASOD} due to their wide-ranging practical applications, such as environmental monitoring, urban planning, military reconnaissance, $etc$. In contrast to common object segmentation in natural scene images (NSIs) \cite{DSP,TIP}, remote sensing objects are more challenging \cite{ORSSD} due to arbitrary orientations, drastic scale changes, and complex backgrounds of targets. Moreover, ORSIs are typically collected by sensors on aircraft or satellites and presented as bird's-eye views. Compared to NSIs, they suffer from unstable imaging conditions and greater susceptibility to interference, which increases the difficulty of the object segmentation task. This means that the designed architecture should pay more attention to how to effectively leverage a variety of significant information.
\begin{figure*}[]
    \centering\includegraphics[width=\textwidth,height=6.43cm]{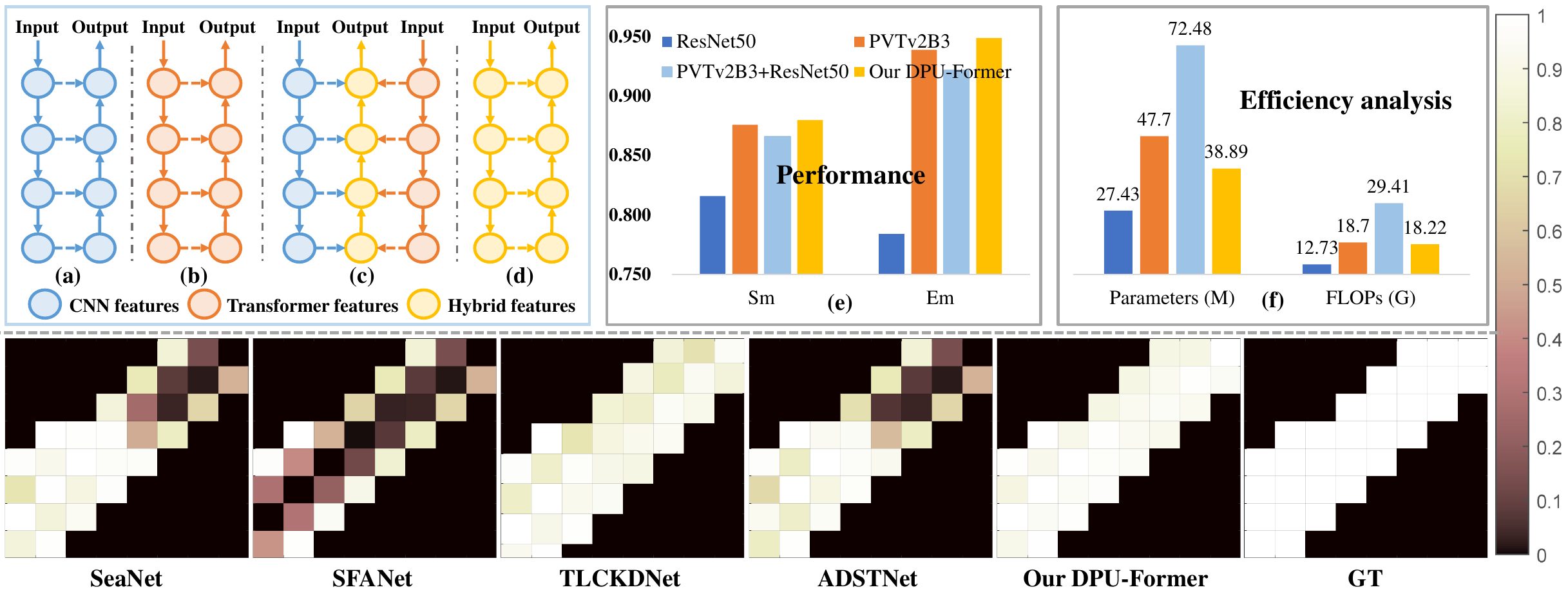}
	\captionsetup{font={small}}
	\caption{\textbf{Top left}: Framework comparison. (a) CNN-based method. (b) Transformer-based method. (c) Hybrid structure-based method. (d) Our DPU-Former method. \textbf{Top right}: Performance and efficiency comparison based on different encoders ($i.e.$, ResNet, PVTv2, ResNet+PVTv2, and our DPU-Former) on the EORSSD dataset. \textbf{Bottom}: Accuracy confusion matrix results.}
	\label{Fig1}
\end{figure*}

Existing ORSIs object segmentation methods are mainly based on data-driven deep learning approaches. As shown in panels (a)-(c) of Fig. \ref{Fig1}, based on differences in encoding features, we classify them into three types, that is, CNN-based, Transformer-based, and hybrid structure-based features. Early, ORSI-based object segmentation methods extract initial features using convolutional encoders ($e.g.$, VGG16 \cite{VGG16} or ResNet50 \cite{ResNet}). These initial features are then refined through various optimization strategies, including multi-scale learning \cite{MFENet,SASOD}, boundary guidance \cite{ERPNet,AESINet}, attention mechanisms \cite{EORSSD,BAttention,RCNet}, and among others. Despite their success, CNN-based methods struggle to capture the relationships between all pixels, limiting the network's ability to achieve global perception due to the inherent constraints of the receptive field. As shown in SeaNet \cite{SeaNet} and SFANet \cite{SFANet} in Fig. \ref{Fig1}, relying solely on local receptive fields can lead to incomplete segmentation results, particularly for remote sensing objects that span a relatively large area in the input ORSIs.

To tackle this limitation, Transformer-based methods are proposed, which utilize self-attention to model long-range dependencies that benefit global understanding. Specifically, GeleNet \cite{GeleNet}, and TLCKDNet \cite{TLKCDNet} adopt a Transformer encoder ($e.g.$,  ViT \cite{ViT} or PVTv2 \cite{Pvt2}) to extract initial features with global information for segmenting remote sensing objects. However, as we know, Transformer-based models lack local inductive bias when processing images \cite{ViT-adapter,GLCONet}, resulting in initial features that are deficient in spatial detail information, which leads to the inferior prediction of clear object boundaries (As depicted in TLCKDNet \cite{TLKCDNet} in Fig. \ref{Fig1}). For ORSIs object segmentation, effectively utilizing global relationships and local details is crucial for accurate results. Recently, hybrid structure-based methods \cite{HFANet,ADSTNet,ASNet} have emerged, which directly combine parallel CNN and Transformer features (as illustrated in Fig. \ref{Fig1} (c)) to obtain features that incorporate both global semantic relationships and local spatial details.

In fact, directly aggregating two types of features in the spatial domain is challenging. As presented in Fig. \ref{Fig1}(e), in our experiments, we observe that when features from different architectures are directly aggregated, their performance, although significantly higher than CNN-based methods, is slightly lower than that of Transformer-based methods. This is unexpected, as the two feature types should theoretically be complementary, yet the experiment shows a discrepancy. The reasons can be attributed to two factors: on the one hand, the internal structure and output paradigm of the Transformer differ significantly from those of the CNN \cite{CT-H,CNN-Transformer}, leading to distinct feature representations and information distributions. On the other hand, these hybrid models use independently trained public encoders, meaning the features are not influenced or adapted to each other during the pre-training process. As a result, there is a lack of correlation between the features. Directly integrating these features may introduce unintended noise, as the two feature types do not fully align in the semantic space, potentially bringing in irrelevant or conflicting information, which can lead to unsatisfactory outcomes (as demonstrated in ADSTNet \cite{ADSTNet} in Fig. \ref{Fig1}). Besides, hybrid structure-based methods often involve many parameters due to the use of two parallel encoders, which increases the complexity of models (as depicted in Fig. \ref{Fig1}(f) ). Consequently, both training and inference become time-consuming and resource-intensive.

To address the above challenges, we propose a novel framework named the Dual-Perspective United Transformer (DPU-Former) for object segmentation in ORSIs. As indicated in Fig. \ref{Fig1}(d), unlike existing ORSIs object segmentation frameworks that rely on the public encoder, DPU-Former utilizes a unified architecture ($i.e.$, Global-local mixed attention, called GLMA) to model local details and global relationships of input images simultaneously, which introduces the Fourier-space merging strategy that projects heterogeneous CNN and Transformer features from the spatial domain to the Fourier space, unifying them into the same frequency band for efficient aggregation. Moreover, the DPU-Former encoder incorporates a gated linear feed-forward network (GLFFN) for enhancing the model's expressive capability. Different from the previous FFN \cite{Pvt2}, our GLFFN further increases diversity of features and selectively filters redundancy information. For the DPU-Former encoder, the dual-perspective simultaneous modeling strategy in GLMA enables the two feature types to enhance their complementarity through iterative adaptation during the pretraining stage. Furthermore, the DPU-Former decoder constructs the adaptive cross-fusion attention and the structural enhancement module, inheriting the benefits of our GLMA, to strengthen and integrate multi-source information from different stages and enhance the structural consistency of objects, thereby improving the accuracy of ORSIs object segmentation.

The proposed DPU-Former model does not rely on public encoders ($e.g.$, ResNet50 \cite{ResNet}, VGG16 \cite{VGG16}, or PVTv2 \cite{Pvt2}) that model data from a single perspective. It extracts complementary information from global and local perspectives, effectively balances global dependencies and local details, and addresses the problem of heterogeneity between CNN and Transformer features. Extensive experimental results on widely-used ORSIs datasets demonstrate the superiority of our DPU-Former method against 16 state-of-the-art (SOTA) models. Additionally, we extend the DPU-Former model to multiple visual scenarios ($i.e.$, natural, camouflage, and medical images), achieving outstanding performance. 

\begin{figure*}[]
	\centering\includegraphics[width=\textwidth,height=5.6cm]{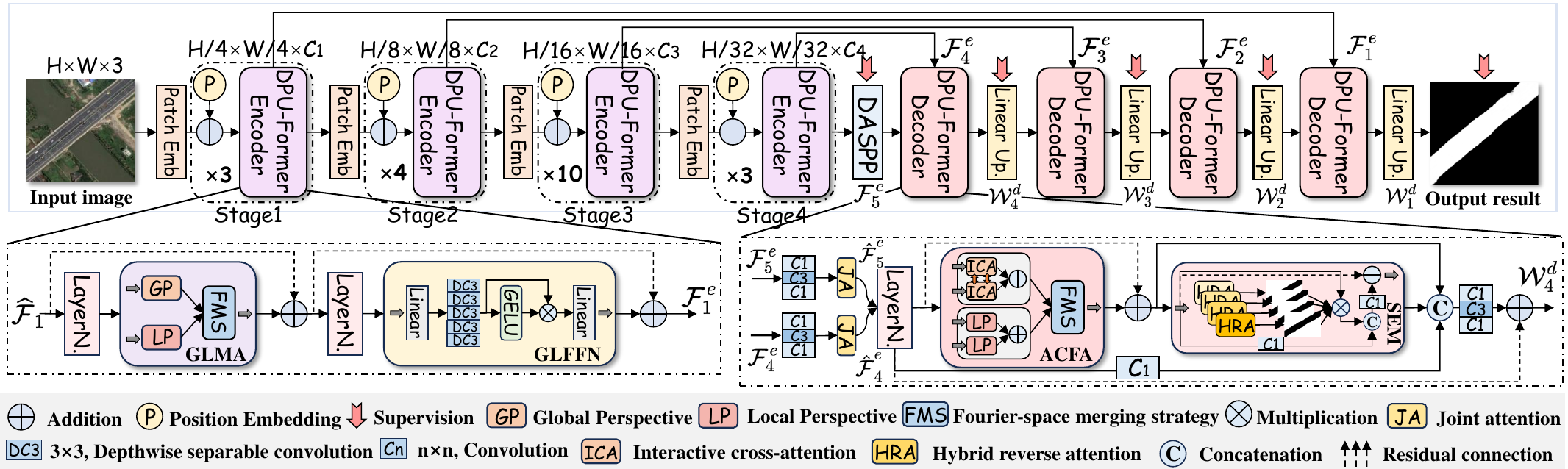}
	\captionsetup{font={small}}
	\caption{Overall architecture of Dual-Perspective United Transformer (DPU-Former), which is divided into the encoder-decoder stage. The DPU-Former encoder includes the global-local mixed attention (GLMA) and the gated linear feed-forward network (GLFFN) to extract initial features with abundant global and local information. The DPU-Former decoder comprises the adaptive cross-fusion attention (ACFA) and the structural enhancement module (SEM) to integrate and enhance multi-level features, thereby generating high-quality representations.}
	\label{Fig2}
\end{figure*}

\section{Related Work}

\textbf{Object Segmentation in ORSIs.} Compared to natural scene images, ORSIs possess characteristics such as irregular topology structures, extreme-scale variations, and imaging interference, making object segmentation tasks for ORSIs particularly challenging. To address these issues, numerous CNN-based methods \cite{ERPNet,SASOD,SFANet} have been proposed, which consistently employ various strategies to optimize the initial features from the CNN encoder for detecting remote sensing objects within scenes. Specifically, LVNet \cite{ORSSD} first constructed an ORSIs dataset and proposed the L-shaped module to perceive the diverse scales and local details of objects. DAFNet \cite{EORSSD} designed the dense attention fluid structure to integrate multi-level attention cues. Next, ERPNet \cite{ERPNet} utilized a shared CNN encoder and two parallel decoders ($i.e.$, edge extraction and feature fusion) to segment remote sensing objects. Similarly, MJRBM \cite{ORSIs-4199}, AESINet \cite{AESINet}, and SeaNet \cite{SeaNet} used edge cues to improve segmentation accuracy. Furthermore, SDNet \cite{SASOD}, and SFANet \cite{SFANet} introduced a series of convolutional modules to increase the diversity. Although the above CNN-based methods attain superior performance, they are limited by the receptive field of convolutional structures, which can lead to only local attribute features in the proposed models. 

Accordingly, researchers attempt to introduce Transformer frameworks \cite{ASTT,Diff-Reg,UDCNet,Nav} with global perception. For example, GeleNet \cite{GeleNet} exploited PVT blocks \cite{Pvt2} to capture global contexts. TLCKDNet \cite{TLKCDNet} adopted the ResT \cite{ResT} as the extractor to provide effective representations. However, Transformer-based models generally suffer from poor handling of local bias, which is not conducive to edge details. Recently, HFANet \cite{HFANet} and ADSTNet \cite{ADSTNet} incorporated CNN and Transformer encoders to obtain initial features with abundant local and global information. However, they overlook the heterogeneity in the CNN and Transformer features and adopt two separate public encoders, increasing the model's burden. Differently, we propose a novel DPU-Former model, which no longer utilizes the public encoders to extract initial features. It is based on our unique architectures to capture local details and global relationships simultaneously and projects two types of features into Fourier space to eliminate their heterogeneity.

\section{Methodology}
The overview framework of our DPU-Former is illustrated in Fig. \ref{Fig2}, consisting of an encoder-decoder architecture. The following DPU-Former encoder, which is composed of a global-local mixed attention (GLMA) and a gated linear feed-forward Network (GLFFN) at each stage, is applied to capture initial encoding features with sufficient local details and global semantics. Additionally, we introduce an adaptive cross-fusion attention (ACFA) and a structural enhancement module (SEM) into the designed DPU-Former decoder, aimed at integrating diverse information at different layers and enhancing the consistency of object structures.

\subsection{DPU-Former Encoder}
For ORSIs object segmentation tasks, a strong encoder is more advantageous for achieving accurate results. However, existing methods rarely consider this important factor, focusing instead on optimizing the initial features provided by public encoders. This leads to an awkward problem where initial features from poor-quality encoders make it difficult to achieve great performance, despite various optimization strategies. Based on this, we design a DPU-Former encoder that simultaneously encodes input images from local and global perspectives to obtain powerful initial features.

\textbf{Global-local mixed attention} enhances the encoder's ability to capture both fine-grained details and broader contexts, which consists of three components ($i.e.$, global perspective, local perspective, and Fourier-space merging strategy). Concretely, given an input image $I\in \mathbb{R}^{H \times W \times 3}$, we first utilize the patch and position embeddings to divide the image into patches with positional information and then use a layer normalization to generate the feature $\check{\mathcal{F}}_{1}$ with a stable input distribution. In the \textbf{global perspective}, we adopt efficient self-attention to model long-range dependencies between all pixels. Technically, we exploit linear layers to generate the $Query$ $(\mathcal{Q}_1)$, $Key$ $(\mathcal{K}_1)$, and $Value$ $(\mathcal{V}_1)$ required for self-attention, which can be defined as:
\begin{equation}
	\begin{split}
		&\mathcal{Q}_1, \mathcal{K}_1, \mathcal{V}_1  =\Theta (\check{\mathcal{F}}_{1}), \Theta (\mathbb{SR}(\check{\mathcal{F}}_{1})),\Theta (\mathbb{SR}(\check{\mathcal{F}}_{1})), \\
	\end{split}
\end{equation}
where $\Theta(\cdot)$ denotes a linear layer, $\mathbb{SR}(\cdot)$ is the spatial reduction \cite{Pvt2} to reduce the complexity. Subsequently, we establish the relationships between all pixels to generate the global feature $\check{\mathcal{F}}^{g}_1$ to focus on the object body information, $i.e.$, $\check{\mathcal{F}} ^{g}_1 = \delta(\frac{\mathcal{Q}_1 \mathcal{K}^{\top }_1 }{\sqrt{d_k} } )\mathcal{V}_1,$ where $\sqrt{d_k}$ presents a scaling factor to prevent values from becoming too large during the dot-product operation, $\delta(\cdot)$ is the Softmax function. Meanwhile, in the \textbf{local perspective}, we adopt a series of well-designed convolutional operations to extract local details, which contain abundant object boundaries. Different from the classic plugin \cite{RFB}, we select the lightweight depthwise separable convolution, which features lower parameters and complexity, and introduce short connections to enhance the correlations between different local features. Specifically, the input feature $\check{\mathcal{F}}_{1}$ is first resized to $\frac{H}{4} \times \frac{W}{4} \times C_1$, and then multi-scale local features are extracted by combining  pointwise convolutions and depthwise separable convolutions, which can be formulated as:
\begin{equation}
	\begin{split}
		&\check{\mathcal{F}}^{l_{k}}_1 = \mathcal{DC}_n\mathcal{C}_1(\check{\mathcal{F}}_{1}+\check{\mathcal{F}}^{l_{k-1}}_1), k=1,2,3\\
	\end{split}
\end{equation}
where $\mathcal{C}_1$ represents a 1 $\times$ 1 pointwise convolution, $\mathcal{DC}_n$ denotes a depthwise separable convolution of $n \times n$ kernels, and we set $n$ to $2k+1$. $+$ is an element-wise addition. Afterward, we aggregate these features to obtain a local feature $\check{\mathcal{F}}^{l}_1$ with abundant spatial details, $i.e.$, $\check{\mathcal{F}}^{l}_1 = \mathcal{C}_1\sum_{k=1}^{3} \check{\mathcal{F}}^{l_k}_1$. 

\begin{figure}[t]
	\centering\includegraphics[width=0.48\textwidth,height=2.9cm]{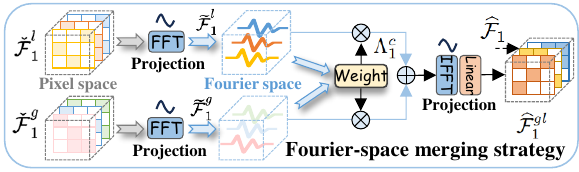}
	\captionsetup{font={small}}
	\caption{Illustration of the Fourier-space merging strategy (FMS).}
	\label{fmsl}
\end{figure}

Although features $\check{\mathcal{F}}^{l}_1$ and $\check{\mathcal{F}}^{g}_1$ are theoretically complementary, the different architectural paradigms of CNN and Transformer lead to the heterogeneity \cite{CT-H,CNN-Transformer} of local and global features in the spatial domain, resulting in direct aggregation not maximizing the advantages of both. Based on this, we propose the \textbf{Fourier-space merging strategy}, which transforms different spatial features into spectral features with the same frequency range, placing them on the same baseline to eliminate the differences in spatial features \cite{FSEL}. Technically, as depicted in Fig. \ref{fmsl}, we employ the Fast Fourier Transform to map the local and global features into the Fourier space, $i.e.$,
\begin{equation}
	\begin{split}
		&\widetilde{\mathcal{F}}^{l}_1,\widetilde{\mathcal{F}}^{g}_1 = \mathbb{FFT}[\check{\mathcal{F}} ^{l}_1],\mathbb{FFT}[\check{\mathcal{F}} ^{g}_1],\\
        &\mathbb{FFT}[\cdot]=\sum_{x=0}^{W-1}\sum_{y=0}^{H-1}f(x,y)e^{-i2\pi (\frac{ux}{W}+\frac{vy}{H})},\\
	\end{split}
\end{equation}
where $\mathbb{FFT}[\cdot]$ is the Fast Fourier Transform, $f(x,y)$ denotes a spatial feature with pixel coordinates $(x, y)$, $i$ is an imaginary unit, $u$ and $v$ present horizontal and vertical indexes in the Fourier space. Furthermore, inspired by the enhancement of spectral features through exploiting a set of weights \cite{AFFT}, we learn a set of composite weights to strengthen the significant information in spectral features. Different from the weight for a single source, our composite weight $\Lambda_1^{c}$ is derived from the features of both $\widetilde{\mathcal{F}}^{l}_1$ and $\widetilde{\mathcal{F}}^{g}_1$, effectively integrating local and global contextual information. $\Lambda_1^{c}$ can be represented as follows:
\begin{equation}
	\begin{split}
		&\Lambda_1^{c}=\varrho (\mathcal{C}_1\mathbb{BR}(\mathcal{C} _1\widetilde{\mathcal{F}}^{l}_1))+\varrho (\mathcal{C}_1\mathbb{BR}(\mathcal{C} _1\widetilde{\mathcal{F}}^{g}_1)), \\
	\end{split}
\end{equation}
where $\varrho(\cdot)$ is the sigmoid function, $\mathbb{BR}(\cdot)$ denotes a batch normalization and a ReLU function. Afterward, two spectral features are adaptively corrected using the obtained composite weight, which integrates local details and global relationships, enhancing the frequency bands within the spectral features to different degrees. Subsequently, we aggregate the optimized two spectral features and convert them back to the original domain through the Inverse Fast Fourier Transform to obtain the powerful feature $\check{\mathcal{F}} ^{gl}_1$, as followed: 
\begin{equation}
	\begin{split}
		&\widehat{\mathcal{F}} ^{gl}_1=\Theta(\mathbb{IFFT} [\Lambda_1^{c}\widetilde{\mathcal{F}}^{l}_1+\Lambda_1^{c}\widetilde{\mathcal{F}}^{g}_1+\widetilde{\mathcal{F}}^{l}_1+\widetilde{\mathcal{F}}^{g}_1])+\widehat{\mathcal{F}}_1, \\
             &\mathbb{IFFT}[\cdot]=\frac{1}{WH}\sum_{u=0}^{W-1}\sum_{v=0}^{H-1}f(u,v)e^{i2\pi(\frac{ux}{W}+\frac{ux}{H} )},\\
	\end{split}
\end{equation}
where $\mathbb{IFFT}[\cdot]$ denotes the Inverse Fast Fourier Transform. 

%Additionally, it incorporates hierarchical depthwise separable convolutions to further enhance feature diversity. 
\textbf{Gated linear feed-forward network} strengthens the model's capacity to learn complex patterns. Unlike the typical FFN \cite{Swin,Pvt2}, our GLFFN introduces a gating mechanism to control the flow of meaningful features and weaken noise. Additionally, it incorporates multiple depthwise separable convolutions to further enhance feature diversity. Technically, for the input feature $\widehat{\mathcal{F}}^{gl}_1$, we first utilize a layer normalization ($\mathbb{LN}(\cdot)$) to generate the feature $\check{\mathcal{F}}_1^{gl}$ ($\check{\mathcal{F}}_1^{gl}=\mathbb{LN}(\widehat{\mathcal{F}}^{gl}_1)$) with a stable distribution. Subsequently, we apply a linear layer and several depthwise separable convolutions to perform linear transformations, generating features $\check{\mathcal{G}} _1^{1}$ and $\check{\mathcal{G}} _1^{2}$. Regarding multiple depthwise separable convolutions, we adopt a hierarchical manner to increase the diversity of features and provide a broad context. The process can be described as follows:
\begin{equation}
	\begin{split}
		&\check{\mathcal{G}} _1^{1}, \check{\mathcal{G}} _1^{2}=\mathcal{DC}_3\Theta(\check{\mathcal{F}}_1^{gl}), [\mathcal{G}_1^{1},\mathcal{G}_1^{2},\mathcal{G}_1^{3},\mathcal{G}_1^{4}]+\check{\mathcal{G}} _1^{1},\\
             &\mathcal{G}_1^{z}=\mathcal{DC}_3(\mathbb{SP}\{\Theta(\check{\mathcal{F}}_1^{gl})\}_z+\mathbb{SP}\{\Theta(\check{\mathcal{F}}_1^{gl})\}_{z-1}),\\
	\end{split}
\end{equation}
where $[\cdot]$ is the concatenation operation, $\mathbb{SP}\{\cdot\}$ represents splitting the feature into four parts based on the number of channels. Note that $z=1, 2, 3, 4$ and $z-1 \ge 1$. Afterward, we utilize the GELU function to perform non-linear activation on the feature $\check{\mathcal{G}} _1^{2}$, and aggregate the activated feature with the feature $\check{\mathcal{G}} _1^{1}$ to control the significant information flow for obtaining high-quality feature $\mathcal{F} _1^e$, as shown in:
\begin{equation}
	\begin{split}
		&\mathcal{F}_1^e=\Theta (\Phi(\check{\mathcal{G}} _1^{2})\otimes \check{\mathcal{G}} _1^{1})+\widehat{\mathcal{F}}^{gl}_1 , \\
	\end{split}
\end{equation}
where $\Phi(\cdot)$ denotes the GELU non-linear activation function, $\otimes$ denotes the element-wise multiplication operation.

In the subsequent stages, $\mathcal{F}_1^e$ serves as input and continues to perform multi-stage feature extraction through the proposed GLMA and GLFFM, ultimately generating three initial features $\mathcal{F}_2^e$, $\mathcal{F}_3^e$, and $\mathcal{F}_4^e$ in our encoder, that is, 
\begin{equation}
	\begin{split}
		&\mathcal{F} _i^{e}=\mathbb{GLFFN}(\mathbb{GLMA}(\mathcal{F} _{i-1}^{e})), i=2,3,4\\
	\end{split}
\end{equation}
where $\mathbb{GLMA}(\cdot)$ and $\mathbb{GLFFN}(\cdot)$ denote the components of GLMA and GLFFM in our DPU-Former encoder.

\subsection{DPU-Former Decoder}
The proposed DPU-Former decoder acts on the initial features $\{\mathcal{F}_i^e\}_{i=1}^4$ from the DPU-Former encoder, with a resolution of $(\frac{H}{2^{i+1}},\frac{W}{2^{i+1}})$ and the number of channels \{64,128,320,512\}, which contains two key components ($i.e.$, adaptive cross-fusion attention and structural enhancement module). Specifically, we first reduce the dimensionality of input initial features to 128 channels using a set of convolution operations, and then adopt a joint attention \cite{CBAM} and a layer normalization to activate positive clues to obtain the feature $\widehat{\mathcal{F}} _i^e$, which is defined as:
\begin{equation}
	\begin{split}
		&\widehat{\mathcal{F}} _i^e =\mathbb{LN}(\mathbb{JA}(\Psi(\mathcal{F}_i^e))), i=4,5\\
	\end{split}
\end{equation}
where $\Psi(\cdot)$ represents a set of convolution operations that contains two 1$\times$1 convolutions and a 3$\times$3 convolution, $\mathbb{JA}(\cdot)$ denotes the joint attention, $\mathcal{F}_5^e$ is obtained by DASPP \cite{DenseASPP} on $\mathcal{F}_4^e$ to provide semantic guidance.

\textbf{Adaptive cross-fusion attention} not only inherits the advantages of our GLMA by optimizing the input features from two perspectives, but also integrates complementary information ($i.e.$, high-level semantics and low-level structures) from different layers of input features. It is worth noting that our ACFA utilizes interactive cross-attention to model long-range dependencies and exchange internal information. Technically, we first employ depthwise separable convolutions and pointwise convolutions to generate the required sets of $Query$, $Key$, and $Value$, and then perform information exchange to obtain the fused global feature $\widehat{\mathcal{W}}_4^g$, as followed:  
\begin{equation}
	\begin{split}
		&\widehat{\mathcal{W}} _4^g = \widehat{\mathcal{F}} _5^{eg}+\widehat{\mathcal{F}} _4^{eg},\\
            &\widehat{\mathcal{F}} _4^{eg},\widehat{\mathcal{F}} _5^{eg}=\delta(\frac{\mathcal{Q}_4 \mathcal{K}^{\top }_4 }{\sqrt{d_k} } )\mathcal{V}_5,\delta(\frac{\mathcal{Q}_5 \mathcal{K}^{\top }_5 }{\sqrt{d_k} } )\mathcal{V}_4.
	\end{split}
\end{equation}
Moreover, we use the same local perspective in our GLMA to capture local information, $i.e.$, $\widehat{\mathcal{W}}_4^l = \mathbb{LP}(\widehat{\mathcal{F}} _5^{e})+\mathbb{LP}(\widehat{\mathcal{F}} _4^{e})$, where $\mathbb{LP}(\cdot)$ denotes the operation of local perspective in the proposed GLMA. Furthermore, the optimized features $\widehat{\mathcal{W}}_4^g$ and $\widehat{\mathcal{W}}_4^l$ are aggregated in Fourier space through the FSM strategy to produce the feature $\widehat{\mathcal{W}} _4^{ac}$, $\widehat{\mathcal{W}} _4^{ac} = \mathbb{FMS}(\widehat{\mathcal{W}}_4^g,\widehat{\mathcal{W}}_4^l)$. 

\textbf{Structural enhancement module} focuses on important region of objects that might have been overlooked and strengthens structural consistency. Specifically, for the input $\widehat{\mathcal{W}}_4^{ac}$, we exploit several atrous convolutions with different filling rates and reverse operations to obtain a hybrid reverse attention map $\mathcal{A}tt_4^h$, that is, $\mathcal{A}tt_4^h= {\sum_{k=1}^{4}}\mathbb{RA}(\mathcal{AC}_3^{2k-1}\mathcal{C}_1 \widehat{\mathcal{W}} _4^{ac})$, where $\mathbb{RA}(\cdot)$ represents reverse attention, $i.e.$, $\mathbb{RA}(\cdot )= -1\otimes \varrho(f(x,y))+1$. $\mathcal{AC}_3^{2k-1}$ denotes the 3$\times$3 atrous convolution with a filling rate of $2k-1$. Furthermore, we embed the hybrid reverse map to enhance structural information to generate the feature $\widehat{\mathcal{W}} _4^{se}$, that is, $\widehat{\mathcal{W}}_4^{se}=\mathcal{C}_1[\mathcal{A}tt_4^h\otimes\mathcal{C}_1 \widehat{\mathcal{W}} _4^{ac},\mathcal{C}_1 \widehat{\mathcal{W}} _4^{ac}]+\widehat{\mathcal{W}} _4^{ac}$.

Finally, we integrate the optimized features $\widehat{\mathcal{W}}_4^{ac}$ and $\widehat{\mathcal{W}}_4^{se}$ to obtain a high-quality representation $\mathcal{W}_4^{d}$ to accurately detect remote sensing targets. It can be formulated as:
\begin{equation}
	\begin{split}
		&\mathcal{W}_4^{d}=\Psi([\widehat{\mathcal{W}} _4^{ac},\widehat{\mathcal{W}}_4^{se},\mathcal{C}_1\widehat{\mathcal{F}} _4^e])+\widehat{\mathcal{F}}_4^e.\\
	\end{split}
\end{equation}
Note that in the subsequent DPU-Former decoder, the feature $\mathcal{W}_{i+1}^{d}$ generated in the previous layer and the same layer feature $\mathcal{F}_{i}^{e}$ from the encoder are used as inputs to gradually aggregate to produce the discriminative feature $\mathcal{W}_{i}^{d}$ through the proposed ACFA and SEM operation.

\subsection{Loss Function}
In our method, we combine the weighted binary cross-entropy (BCE) and the weighted intersection over union (IoU) functions to supervise the training model, as followed:
\begin{equation}
	\begin{split}
		&\mathcal{L}_{com}=\sum_{i=1}^{5}\frac{1}{2^{i-1}}  (\mathcal{L}_{IoU}^{W}(\mathcal{P}_i,G_t)+\mathcal{L}_{BCE}^{W}(\mathcal{P}_i,G_t)),\\
	\end{split}
\end{equation}
where $\mathcal{L}_{IoU}^{W}$ and $\mathcal{L}_{BCE}^{W}$ are the weighted IoU and BCE loss functions. $\mathcal{P}_i$ represents the feature $\mathcal{W}_{i}^{d}$ after being reduced to one channel, and $G_t$ denotes the ground truth. Note that $\mathcal{W}_{5}^{d}$ is the feature $\mathcal{F}_5^e$ from the DASPP \cite{DenseASPP}.

\begin{table*}[]
\centering
\setlength{\tabcolsep}{6pt}
\renewcommand{\arraystretch}{0.8}
\resizebox*{0.95\textwidth}{59mm}{
\begin{tabular}{c|ccccc|ccccc|ccccc}
\hline \hline 
\multirow{2}{*}{\textbf{Methods}} & \multicolumn{5}{c|}{\textbf{ORSIs-4199 (2199 images)}}         & \multicolumn{5}{c|}{\textbf{EORSSD (600 images)}}           & \multicolumn{5}{c}{\textbf{ORSSD (200 images)}}             \\
                       & \cellcolor{cyan!10}$\mathcal{F}_{m}^{w}$$\uparrow$   & \cellcolor{cyan!10}$\mathcal{F}_{m}^{m}$$\uparrow$   & \cellcolor{cyan!10}$S_{m}$$\uparrow$    & \cellcolor{cyan!10}$E_{m}$$\uparrow$    & \cellcolor{cyan!10}$\mathcal{M}$$\downarrow$   & \cellcolor{cyan!10}$\mathcal{F}_{m}^{w}$$\uparrow$   & \cellcolor{cyan!10}$\mathcal{F}_{m}^{m}$$\uparrow$   & \cellcolor{cyan!10}$S_{m}$$\uparrow$    & \cellcolor{cyan!10}$E_{m}$$\uparrow$    & \cellcolor{cyan!10}$\mathcal{M}$$\downarrow$   & \cellcolor{cyan!10}$\mathcal{F}_{m}^{w}$$\uparrow$   & \cellcolor{cyan!10}$\mathcal{F}_{m}^{m}$$\uparrow$   & \cellcolor{cyan!10}$S_{m}$$\uparrow$    & \cellcolor{cyan!10}$E_{m}$$\uparrow$    & \cellcolor{cyan!10}$\mathcal{M}$$\downarrow$   \\ \hline \hline
                       
PA-KRN$_{21}$                  & 0.811 & 0.859 & 0.843 & 0.917 & 0.038 & 0.839 & 0.875 & 0.880 & 0.927 & 0.010 & 0.868 & 0.896 & 0.915 & 0.941 & 0.014 \\ 
VST$_{21}$                     & 0.835 & 0.883 & 0.873 & 0.907 & 0.028 & 0.819 & 0.881 & 0.883 & 0.879 & 0.007 & 0.872 & 0.917 & 0.927 & 0.940 & 0.009 \\ 
DAFNet$_{21}$                  & -     & -     & -     & -     & -     & 0.783 & 0.867 & 0.883 & 0.815 & \textbf{\color{red}0.006} & 0.844 & 0.903 & 0.912 & 0.920 & 0.011 \\ 
ERPNet$_{22}$                  & 0.816 & 0.872 & 0.861 & 0.904 & 0.036 & 0.825 & 0.877 & 0.881 & 0.923 & 0.009 & 0.864 & 0.904 & 0.915 & 0.953 & 0.014 \\ 
EMFINet$_{22}$                 & 0.830 & 0.872 & 0.861 & 0.914 & 0.033 & 0.849 & 0.882 & 0.889 & 0.950 & 0.008 & 0.881 & 0.907 & 0.927 & 0.966 & 0.011 \\ 
CorrNet$_{22}$                 & 0.828 & 0.873 & 0.856 & 0.930 & 0.037 & 0.862 & 0.891 & 0.889 & 0.960 & 0.008 & 0.896 & 0.917 & 0.929 & 0.976 & 0.010 \\ 
HFANet$_{22}$                  & 0.845 & 0.882 & 0.870 & 0.919 & 0.031 & 0.871 & 0.895 & 0.896 & 0.934 & 0.007 & 0.894 & 0.917 & 0.930 & 0.952 & 0.009 \\ 
MJRBM$_{22}$                   & 0.806 & 0.867 & 0.853 & 0.909 & 0.037 & 0.813 & 0.877 & 0.879 & 0.890 & 0.009 & 0.844 & 0.893 & 0.910 & 0.934 & 0.016 \\ 
SeaNet$_{23}$                  & 0.842 & 0.878 & 0.866 & 0.939 & 0.031 & 0.850 & 0.874 & 0.884 & 0.961 & 0.007 & 0.873 & 0.899 & 0.917 & 0.971 & 0.011 \\ 
AESINet$_{23}$                 & -     & -     & -     & -     & -     & 0.853 & 0.886 & 0.896 & 0.928 & \textbf{\color{red}0.006} & 0.895 & 0.919 & \textbf{\color{red}0.935} & 0.954 & 0.009 \\ 
SARL$_{23}$                    & 0.842 & 0.880 & 0.868 & 0.919 & 0.031 & 0.851 & 0.880 & 0.887 & 0.926 & 0.007 & 0.874 & 0.906 & 0.923 & 0.944 & 0.011 \\ 
SDNet$_{23}$                   & 0.845 & 0.884 & 0.871 & 0.922 & 0.030 & 0.870 & 0.893 & 0.895 & 0.931 & 0.007 & 0.889 & 0.913 & 0.924 & 0.949 & 0.010 \\ 
ICON$_{23}$                    & 0.852 & 0.882 & 0.869 & 0.944 & 0.028 & 0.843 & 0.871 & 0.882 & 0.922 & 0.007 & 0.867 & 0.904 & 0.916 & 0.943 & 0.012 \\ 
ADSTNet$_{24}$                 & 0.848 & 0.883 & 0.865 & 0.941 & 0.032 & 0.869 & 0.887 & 0.891 & \textbf{\color{red}0.969} & 0.007 & 0.899 & 0.918 & 0.930 & 0.982 & 0.009 \\ 
TLCKDNet$_{24}$                & -     & -     & -     & -     & -     & 0.857 & 0.891 & 0.896 & 0.921 & \textbf{\color{red}0.006} & 0.892 & 0.920 & 0.931 & 0.950 & 0.008 \\ 
SFANet$_{24}$                  & 0.850 & 0.883 & 0.870 & 0.942 & 0.029 & 0.872 & 0.893 & 0.896 & 0.968 & \textbf{\color{red}0.006} & 0.907 & 0.923 & \textbf{\color{red}0.935} & 0.980 & 0.008 \\ \hdashline
\rowcolor{magenta!13}Ours                    & \textbf{\color{red}0.870} & \textbf{\color{red}0.896} & \textbf{\color{red}0.877} & \textbf{\color{red}0.953} & \textbf{\color{red}0.026} & \textbf{\color{red}0.881} & \textbf{\color{red}0.900} & \textbf{\color{red}0.901} & 0.966 & \textbf{\color{red}0.006} & \textbf{\color{red}0.916} & \textbf{\color{red}0.929} & 0.931 & \textbf{\color{red}0.984} & \textbf{\color{red}0.006} \\ \hline \hline
\end{tabular}}
\caption{Quantitative results on three widely-used ORSIs object segmentation datasets. The best result is marked in \textbf{\color{red}blod}. The symbols ``$\uparrow$/$\downarrow$'' denote that a higher/lower value is better. }
\label{Table-QR}
\end{table*}
\begin{table*}[t]
\centering
\resizebox*{0.92\textwidth}{10mm}{
\begin{tabular}{c|cccccccccc}
\hline
              \rowcolor{magenta!15}& PA-KRN$_{21}$ & DFANet$_{21}$ & MJRBM$_{22}$ & ERPNet$_{22}$ & EMFINet$_{22}$ & ACCoNet$_{22}$ & SDNet$_{23}$ & ADSTNet$_{24}$ & TLCKDNet$_{24}$ & Ours  \\ \hline \hline
\textbf{Parameters(M)} $\downarrow$ & 141.06 & \textbf{\color{red}29.35}  & 43.54 & 56.48  & 95.09   & 102.55   & 61.50 & 62.09   & 52.09    & 44.20 \\ 
\textbf{FLOPs(G)} $\downarrow$     & 630.17 & 75.37  & 95.81 & 215.45 & 334.52  & 348.81  & 73.23 & 40.53   & 59.98    & \textbf{\color{red}32.51} \\ \hline 
\end{tabular}}
\caption{ Efficiency analysis of parameters and FLOPs. }
\label{table-PF}
\end{table*}

\section{Experiments}
\subsection{Experimental Settings}
\textbf{Datasets and Evaluation Metrics.} We conduct experiments on three ORSIs datsets, including ORSSD \cite{ORSSD}, EORSSD \cite{EORSSD}, and ORSIs-4199 \cite{ORSIs-4199}. ORSSD comprises 600 images for training and 200 images for testing. EORSSD contains 1400 training images and 600 testing images. ORSIs-4199 is the most challenging dataset and consists of 2000 images for training and 2199 images for testing. Furthermore, we adopt several evaluation metrics to evaluate our model, including the Weight F-measure ($F_{m}^{w}$), Maximum F-measure ($F_{m}^{m}$), S-measure ($S_m$)\nocite{Sm}, and E-measure ($E_m$), and Mean Absolute Error ($\mathcal{M}$).

%We follow the strategy of existing approaches \cite{SRAL,SFANet} by performing independent training and testing for each dataset. In addition, we adopt several evaluation metrics to evaluate our model, including the Weight F-measure ($F_{m}^{w}$), Maximum F-measure ($F_{m}^{m}$), S-measure ($S_m$)\nocite{Sm}, and E-measure ($E_m$), and Mean Absolute Error ($\mathcal{M}$).

\textbf{Implementation details.} We leverage the PyTorch framework to implement our DPU-Former, training it on a setup of four NVIDIA GTX 4090 GPUs. Similarly to the typical ResNet \cite{ResNet} encoder, the DPU-Former encoder is pre-trained on ImageNet-1k \cite{imagenet}. In the ORSIs object segmentation task, we employ the Adam optimizer with an initial learning rate set to 1e-4 and apply weight decay at a rate of 0.1 every 40 epochs. The batch size is set to 48. During training, the input images are uniformly cropped to 352$\times$ 352, and the whole process takes 80 epochs. 
Following \cite{SFANet}, we apply data augmentation techniques ($e.g.$, horizontal flipping and rotation) to improve the model's adaptability to remote sensing targets.
\begin{figure}[t]
	\centering\includegraphics[width=0.48\textwidth,height=4.75cm]{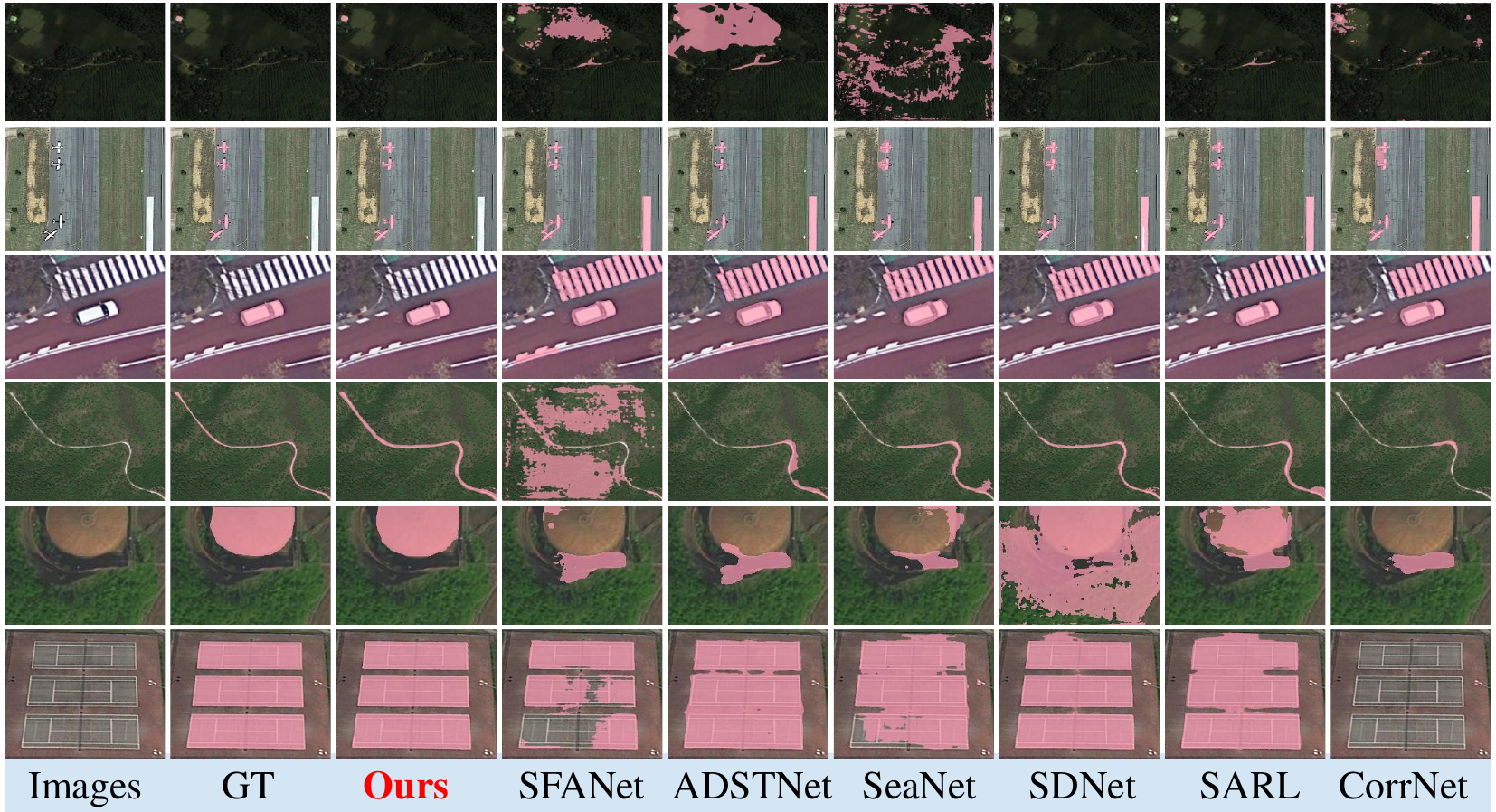}
	\captionsetup{font={small}, justification=raggedright}
	\caption{Visual results of DPU-Former and existing methods.}
	\label{Fig-visual_results}
\end{figure}
\begin{table}[t]
\centering
\setlength{\tabcolsep}{3pt}
\resizebox{0.48\textwidth}{17mm}{
\begin{tabular}{c|cccc|cc|ccc|ccc}
\hline \hline
\multirow{3}{*}{\textbf{Num.}} & \multicolumn{4}{c|}{\textbf{Encoder}}                                & \multicolumn{2}{c|}{\textbf{Decoder}}                 & \multicolumn{3}{c|}{\multirow{2}{*}{\textbf{EORSSD}}} & \multicolumn{3}{c}{\multirow{2}{*}{\textbf{ORSIs-4199}}} \\ \cline{2-7}
                      & \multicolumn{3}{c|}{\textbf{GLMA}}          & \multirow{2}{*}{\textbf{GLFFN}} & \multirow{2}{*}{\textbf{ACFA}} & \multirow{2}{*}{\textbf{SEM}} & \multicolumn{3}{c|}{}                        & \multicolumn{3}{c}{}                       \\
                      & \textbf{GP} & \textbf{LP} & \multicolumn{1}{c|}{\textbf{FMS}} &                        &                       &                      &  \cellcolor{cyan!10} $\mathcal{F}_{m}^{w}$$\uparrow$           & \cellcolor{cyan!10}$S_m$$\uparrow$            & \cellcolor{cyan!10}$E_m$$\uparrow$           & \cellcolor{cyan!10}$\mathcal{F}_{m}^{w}$$\uparrow$          & \cellcolor{cyan!10}$S_m$$\uparrow$           & \cellcolor{cyan!10}$E_m$$\uparrow$           \\ \hline \hline
(a)                   & $\checkmark$  &    & \multicolumn{1}{c|}{}    &                        &                       &                      & 0.791         & 0.856         & 0.906        & 0.815        & 0.839        & 0.917        \\ 
(b)                   & $\checkmark$  &    & \multicolumn{1}{c|}{}    & $\checkmark$                      &                       &                      & 0.819         & 0.862         & 0.932        & 0.841        & 0.858        & 0.935        \\ 
(c)                   & $\checkmark$  & $\checkmark$  & \multicolumn{1}{c|}{}    & $\checkmark$                      &                       &                      & 0.842         & 0.870         & 0.946        & 0.853        & 0.863        & 0.943        \\ 
(d)                   & $\checkmark$  & $\checkmark$  & \multicolumn{1}{c|}{$\checkmark$}   & $\checkmark$                      &                       &                      & 0.849         & 0.878         & 0.949        & 0.855        & 0.868        & 0.945        \\ 
(e)                   & $\checkmark$  & $\checkmark$  & \multicolumn{1}{c|}{$\checkmark$}   &                      &                       &                      & 0.842         & 0.876         & 0.948        & 0.849        & 0.860        & 0.931        \\ 
(f)                   & $\checkmark$  & $\checkmark$  & \multicolumn{1}{c|}{$\checkmark$}   & $\checkmark$                      & $\checkmark$                     &                      & 0.864         & 0.897         & 0.952        & 0.863        & \textbf{\color{red}0.877}        & 0.949        \\ 
(g)                   & $\checkmark$  & $\checkmark$  & \multicolumn{1}{c|}{$\checkmark$}   & $\checkmark$                      &                       & $\checkmark$                    & 0.872         & 0.899         & 0.959        & 0.863        & 0.873        & 0.949        \\ \hdashline
\rowcolor{magenta!13}(h)                   & $\checkmark$  & $\checkmark$  & \multicolumn{1}{c|}{$\checkmark$}   & $\checkmark$                      & $\checkmark$                     & $\checkmark$                    & \textbf{\color{red}0.881}         & \textbf{\color{red}0.901}         & \textbf{\color{red}0.966}        & \textbf{\color{red}0.870}        & \textbf{\color{red}0.877}        & \textbf{\color{red}0.953}        \\ \hline \hline
\end{tabular}}

\caption{Ablation analysis of all components in our DPU-Former.}
\label{TABLE-AB}
\end{table}
\begin{figure}[t]
	\centering\includegraphics[width=0.48\textwidth,height=3.2cm]{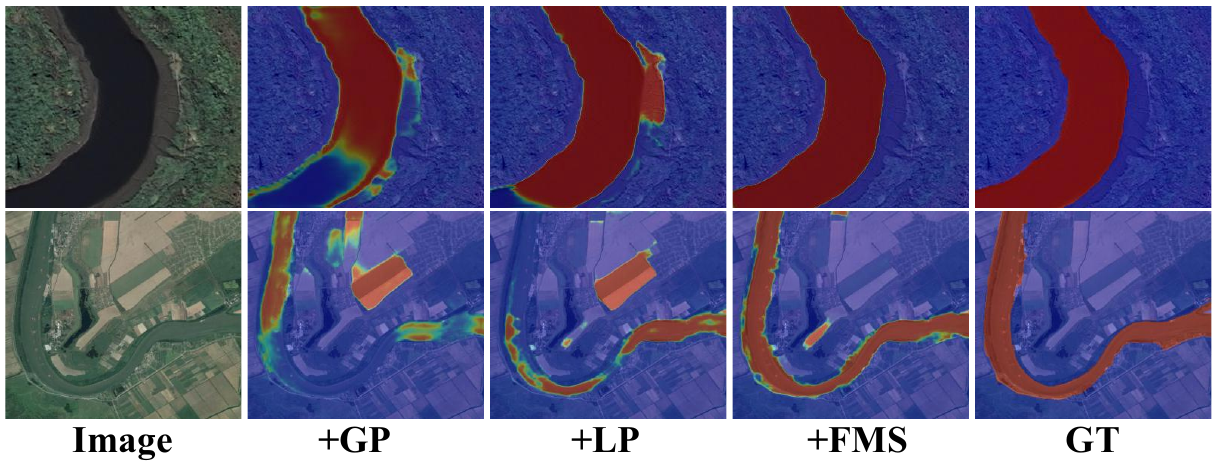}
	\captionsetup{font={small}}
	\caption{Heatmap visualization of the DPU-Former encoder.}
	\label{encoder_visual}
\end{figure}
\begin{figure*}[]
	\centering\includegraphics[width=\textwidth,height=3cm]{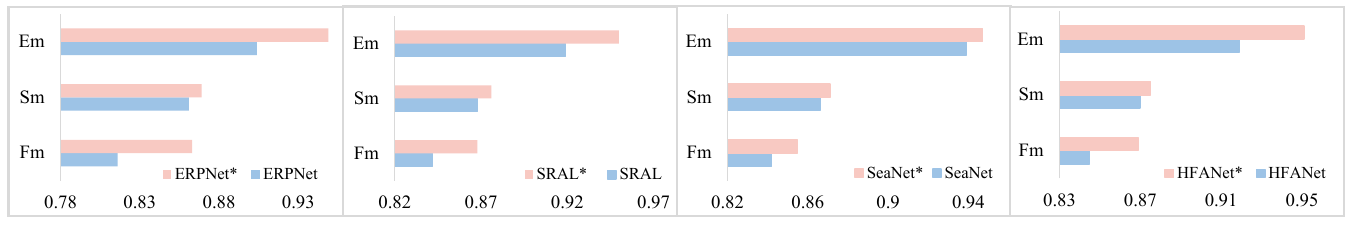}
	\captionsetup{font={small}}
	\caption{Plug-and-play results in the ORSIs-4199 dataset. $^{\ast}$ represents extracting initial features using our DPU-Former encoder.}
	\label{jicha}
\end{figure*}

\subsection{Comparison with the State-of-the-Arts}
We compare the DPU-Former with 16 ORSIs segmentation methods, including PA-KRN\cite{PAKRN}, VST \cite{VST}, DAFNet \cite{EORSSD}, ERPNet \cite{ERPNet}, EMFINet \cite{EMFINet}, CorrNet \cite{CoorNet}, HFANet \cite{HFANet}, MJRBM \cite{ORSIs-4199}, SeaNet \cite{SeaNet}, AESINet \cite{AESINet}, SRAL \cite{SRAL}, SDNet \cite{SASOD}, ICON \cite{ICON}, ADSTNet \cite{ADSTNet}, TLCKDNet \cite{TLKCDNet}, and SFANet \cite{SFANet}. To ensure a fair comparison, all results are provided either directly from the authors or from available open-source code. 

\textbf{Quantitative results.} Table \ref{Table-QR} lists the performance of our DPU-Former against the 16 comparison methods.  Specifically, our $F_{m}^{w}$ scores surpass the second-best method by a clear margin of 1.03\% and 1.10\% in the EORSSD \cite{EORSSD} and ORSSD \cite{ORSSD} datasets, respectively. In the extremely challenging ORSIs-4199 \cite{ORSIs-4199} dataset, our DPU-Former is better than the second-best method in terms of $F_{m}^{w}$ (0.870 $vs.$ 0.852), $F_{m}^{m}$ (0.896 $vs.$ 0.884), $S_m$ (0.877 $vs.$ 0.873), $E_m$ (0.953 $vs.$ 0.944), and $\mathcal{M}$ (0.026 $vs.$ 0.028). Furthermore, we provide the parameters and FLOPs of the model in Table \ref{table-PF}. To ensure fairness, we set the input image size for all methods to 352 $\times$ 352. From Table \ref{table-PF}, it is evident that our model is highly competitive in terms of both the parameters and FLOPs. These quantitative results demonstrate that our DPU-Former method achieves superior accuracy and efficiency.

\textbf{Qualitative results.} Fig. \ref{Fig-visual_results} gives the visual prediction results of the proposed DPU-Former and existing ORSIs segmentation methods under different remote sensing scenes, including building ($1^{st}$ and $5^{th}$ rows), aircraft ($2^{nd}$ row), motorcar ($3^{rd}$ row), river ($4^{th}$ row), and court ($6^{th}$ row). As depicted in Fig. \ref{Fig-visual_results}, our DPU-Former model outperforms the previous SOTA methods ($e.g.$, SFANet \cite{SFANet} and ADSTNet \cite{ADSTNet}) in the processing of remote sensing objects with various types and sizes.

\subsection{Ablation Study}

We conduct extensive ablation experiments to analyze the effectiveness of all proposed components and strategies on two public ORSIs object segmentation datasets.

\begin{table}[]
\centering
\setlength{\tabcolsep}{4pt}
\resizebox{0.48\textwidth}{15mm}{
\begin{tabular}{c|c|c|c|ccc|ccc}
\hline \hline
\multirow{2}{*}{\textbf{Num.}} & \multirow{2}{*}{\textbf{Encoder}} & \multirow{2}{*}{\begin{tabular}[c]{@{}c@{}}\textbf{Paramers}\\ (M)\end{tabular}} & \multirow{2}{*}{\begin{tabular}[c]{@{}c@{}}\textbf{FLOPs}\\ (G)\end{tabular}} & \multicolumn{3}{c|}{\textbf{EORSSD}} & \multicolumn{3}{c}{\textbf{ORSIs-4199}} \\
                      &                          &                                                                         &                                                                      &  \cellcolor{cyan!10} $\mathcal{F}_{m}^{w}$$\uparrow$           & \cellcolor{cyan!10}$S_m$$\uparrow$            & \cellcolor{cyan!10}$E_m$$\uparrow$           & \cellcolor{cyan!10}$\mathcal{F}_{m}^{w}$$\uparrow$          & \cellcolor{cyan!10}$S_m$$\uparrow$           & \cellcolor{cyan!10}$E_m$$\uparrow$      \\ \hline \hline
(a)                   & ResNet50                 & {27.43}                                                                   & 12.73                                                                & 0.667   & 0.816   & 0.784   & 0.827    & 0.858   & 0.920   \\ 
(b)                   & Res2Net                  & 27.59                                                                   & 13.75                                                                & 0.833   & 0.865   & 0.919   & 0.823    & 0.854   & 0.923   \\ 
(c)                   & VGG16                    & \textbf{\color{red}17.74}                                                                   & 47.05                                                                & 0.821   & 0.857   & 0.902   & 0.820    & 0.853   & 0.913   \\ 
(d)                   & Swin-S                   & 51.91                                                                   & \textbf{\color{red}9.43}                                                                 & 0.796   & 0.861   & 0.931   & 0.836    & 0.853   & 0.937   \\ 
(e)                   & PVTv2-B3                  & 47.70                                                                   & 18.70                                                                & \textbf{\color{red}0.857}   & 0.876   & 0.939   & 0.832    & \textbf{\color{red}0.871}   & 0.941   \\ 
(f)                   & Pv2-B3+R50                  & 72.48                                                                   & 29.41                                                                & {0.834}   & 0.866   & 0.922   & 0.846    & {0.866}   & 0.939   \\ \hdashline

\rowcolor{magenta!13}(g)                   & DPU-Former                      & 38.89                                                                   & 18.22                                                                & 0.849   & \textbf{\color{red}0.878}   & \textbf{\color{red}0.949}   & \textbf{\color{red}0.855}    & 0.868   & \textbf{\color{red}0.945}   \\ \hline \hline
\end{tabular}}
\caption{ Ablation analysis of different encoders.}
\label{Table-backa}
\end{table}

\textbf{Effectiveness of the DPU-Former encoder.} In the DPU-Former encoder, the key components are ``GLMA'' and ``GLFFN''. We begin by assessing the influence of different perspectives in our ``GLMA''. From Table \ref{TABLE-AB}, embedding the local perspective (Tab. \ref{TABLE-AB}(c)) into the global perspective (Tab. \ref{TABLE-AB}(b)) significantly improves performance, which is attributed to the increase of local details in input features. Moreover, the Fourier-space merging strategy (Tab. \ref{TABLE-AB}(d)) outperforms element-wise addition (Tab. \ref{TABLE-AB}(c)) from the spatial domain because the representation of unified frequencies in the Fourier domain reduces the impact of differences in features. The visual heatmaps for each stage is shown in Fig. \ref{encoder_visual}. Furthermore, we study the importance of our GLFFN by removing them in experiments. As shown in Table \ref{TABLE-AB} (a) and (e), we can see that ``GLFFN'' is meaningful for the designed DPU-Former encoder due to its powerful linear enhancement and redundancy information filtering. Furthermore, we compare our DPU-former encoder with existing classic encoders ($i.e.$, VGG16 \cite{VGG16}, ResNet50 \cite{ResNet}, Res2Net \cite{Res2Net}, Swin-small \cite{Swin}, PVTv2-b3 \cite{Pvt2}) in Table \ref{Table-backa}. Additionally, as depicted in Fig. \ref{jicha}, we perform plug-and-play studies on the DPU-Former encoder using the four existing methods. Our DPU-Former encoder is highly competitive in the performance and efficiency of segmenting objects.

\textbf{Effectiveness of the DPU-Former decoder.} The DPU-former decoder inherits the advantages of retaining local details and capturing long-range dependencies and introduces cross-fusion and structural reinforcement. We conduct experimental analysis on the proposed ``ACFA'' and ``SEM'' of the proposed DPU-Former decoder. As shown in Table \ref{TABLE-AB}, the individual ACFA (Tab. \ref{TABLE-AB}(f)) and SEM (Tab. \ref{TABLE-AB}(g)) surpass the FPN \cite{FPN} decoder (Tab. \ref{TABLE-AB}(d)), with a notable performance improvement of 2.16\% and 2.39\% respectively in the $\mathcal{S}_{m}$ evaluation metric on EORSSD dataset. Furthermore, the best result (Tab. \ref{TABLE-AB}(h)) can be obtained by combining our ``ACFA'' and ``SEM'', demonstrating the effectiveness and compatibility of these components.

\begin{figure}[]
	\centering\includegraphics[width=0.48\textwidth,height=4cm]{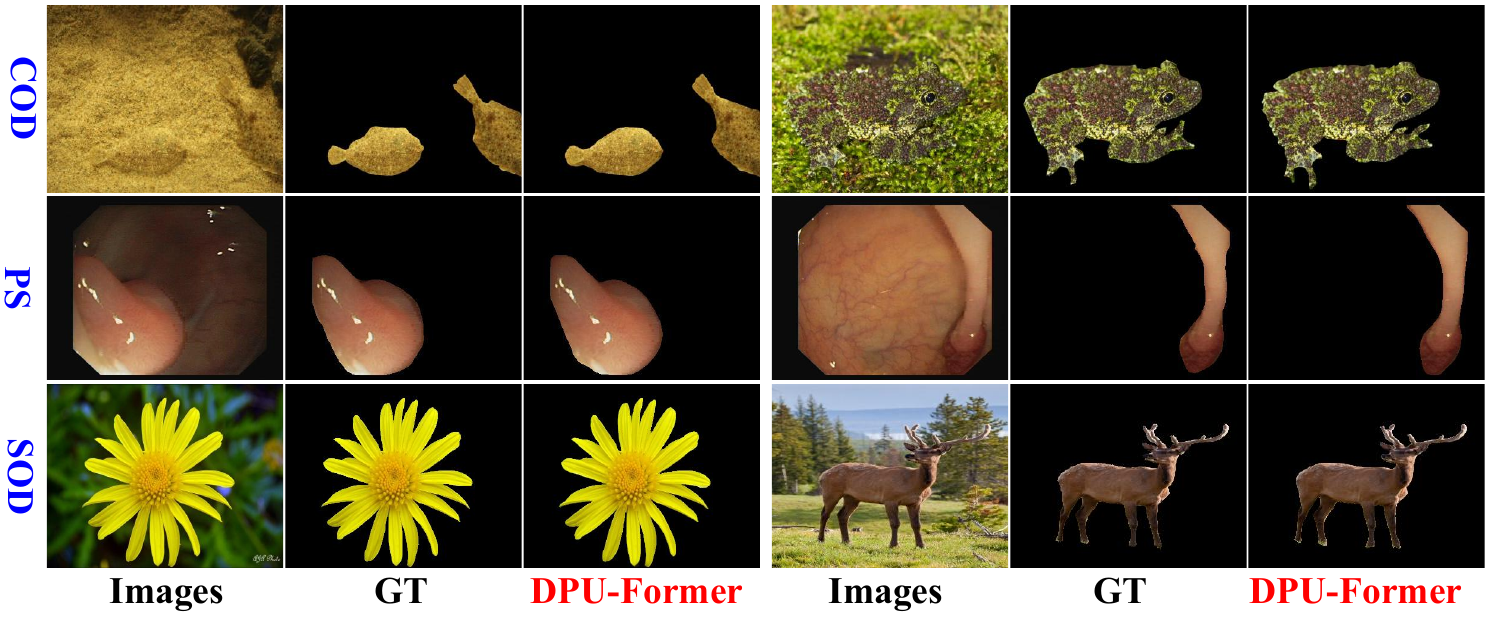}
	\captionsetup{font={small}}
	\caption{Visual segmentation results of the expanded application.}
	\label{expanded}
\end{figure}

\subsection{Extended Applications}
Beyond optical remote sensing images, we also evaluate the generalizability of our DPU-Former model on natural scene images, camouflage images, and medical images. Specifically, we apply the proposed DPU-Former model to three object segmentation tasks, including salient object detection (SOD), camouflage object detection (COD), and polyp segmentation (PS). From Fig. \ref{expanded}, the proposed DPU-Former method achieves outstanding segmentation accuracy, attributed to the cooperation of both local and global perspectives. More details in \textbf{{supplementary materials}}.

\section{Conclusions}
In this paper, we propose the DPU-Former model, a novel framework for the ORSIs object segmentation task. It obtains mutually beneficial features from both local and global perspectives and introduces the FMS to eliminate heterogeneity. Moreover, we present the GLFFN to enhance expression ability. Additionally, within the decoder, we construct the ACFA and SEM to integrate and strengthen multilevel features. As a result, the proposed DPU-Former method achieves outstanding performance on multiple object segmentation datasets.

\section*{Acknowledgments} This work was supported in part by the National Science Fund of China (No. 62276135, 62176124, and 62361166670).

\bibliographystyle{named}
\bibliography{ijcai25}

\end{document}